\documentclass[10pt, conference, compsocconf]{IEEEtran}

\usepackage{epsfig}
\usepackage[caption=false,font=footnotesize]{subfig}
\usepackage{graphicx}
\usepackage{stfloats}
\usepackage{fixltx2e}
\usepackage{cite}
\usepackage{placeins}
\usepackage{amsmath}

\begin{document}

\title{Point Pair Feature based Object Detection for Random Bin Picking}

\author{\IEEEauthorblockN{Wim Abbeloos and Toon Goedem\'e}
\IEEEauthorblockA{KU Leuven, Department of Electrical Engineering, EAVISE\\ Leuven, Belgium\\Email: wim.abbeloos@kuleuven.be, toon.goedeme@kuleuven.be}}

\maketitle

\begin{abstract}
Point pair features are a popular representation for free form 3D object detection and pose estimation.  In this paper, their performance in an industrial random bin picking context is investigated.  A new method to generate representative synthetic datasets is proposed.  This allows to investigate the influence of a high degree of clutter and the presence of self similar features, which are typical to our application.  We provide an overview of solutions proposed in literature and discuss their strengths and weaknesses.  A simple heuristic method to drastically reduce the computational complexity is introduced, which results in improved robustness, speed and accuracy compared to the naive approach.

\end{abstract}

\section{Introduction}
\label{Introduction}
\begin{figure}
	\centering
	\includegraphics[width=0.3\textwidth]{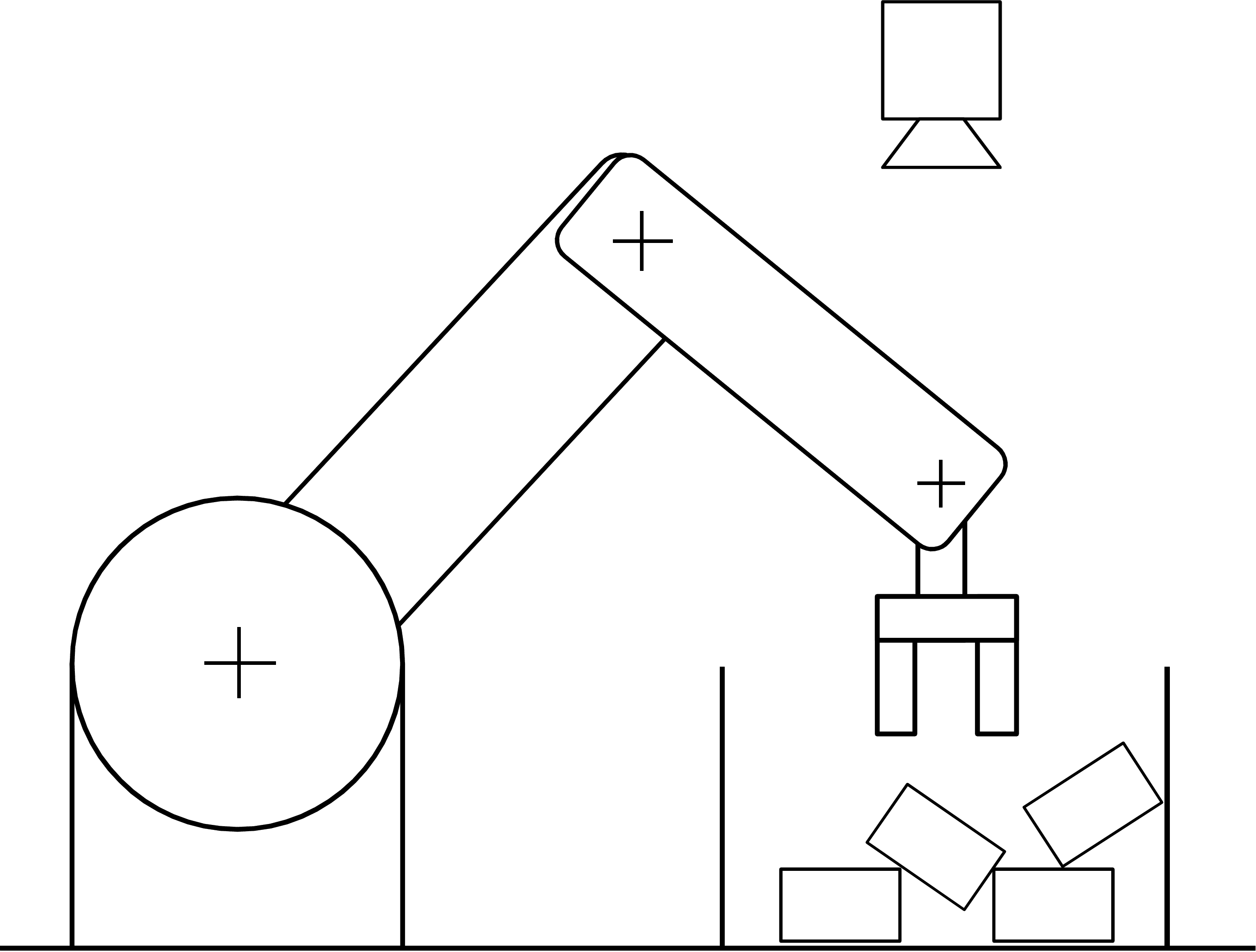}
\caption{A typical bin picking setup consists of some kind of 3D sensor mounted above a bin filled with randomly stacked parts.}
	\label{fig:RBP}
\end{figure}

\begin{figure}
	\centering
	\includegraphics[width=0.3\textwidth]{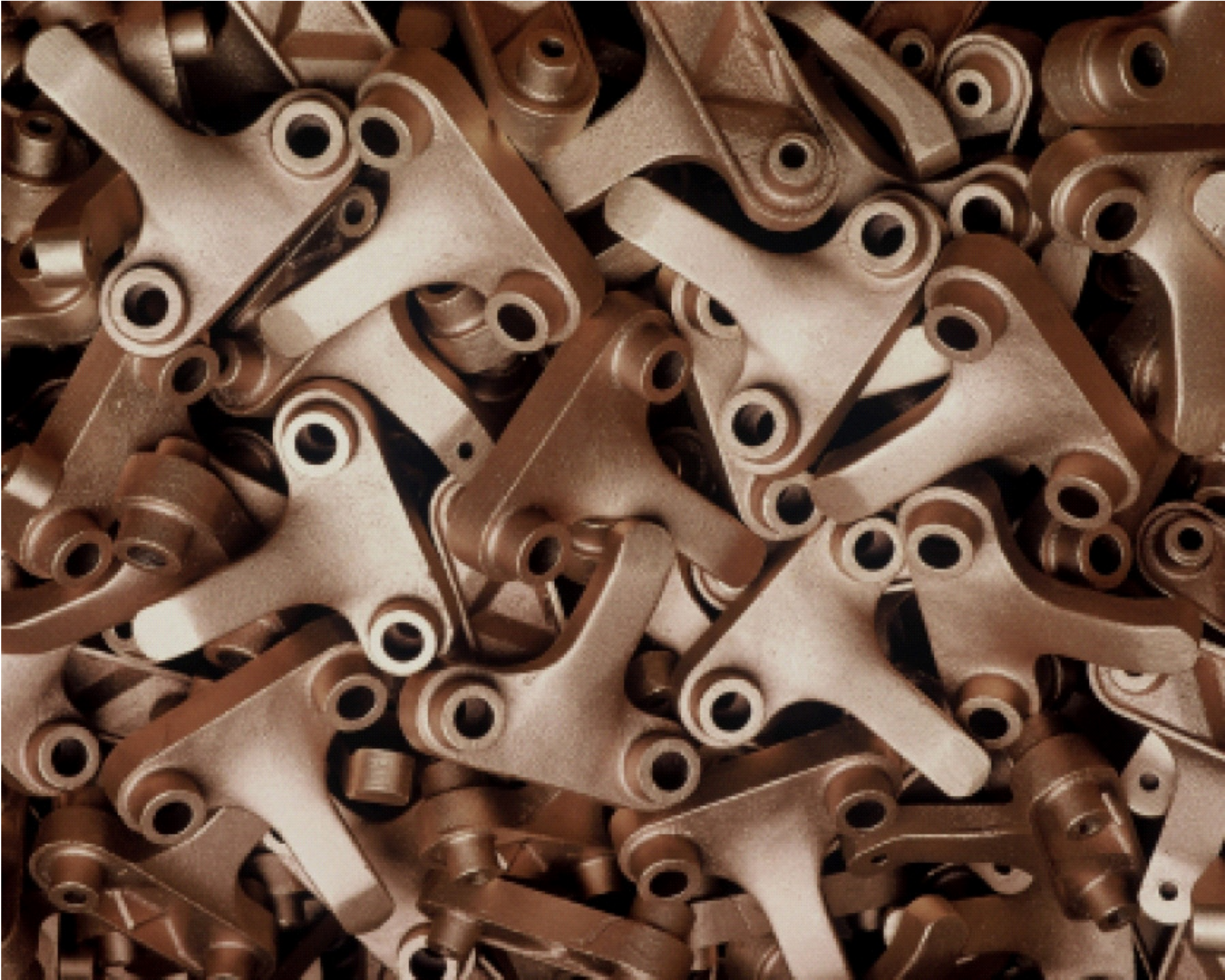}
	\caption{Objects supplied in bulk are stacked randomly in a bin.  An object detection and pose estimation algorithm is needed to determine the six degree of freedom pose of a pickable object, so that it can be grasped by the robot.}
	\label{fig:parts}
\end{figure}

The automatic handling of objects by industrial robots is common practice.  However, if the object's pose is unknown beforehand, it needs to be measured. Indeed, vision-guided industrial robots are one of the key ingredients of state-of-the-art manufacturing processes. Everyone is aware of the fact that future production processes will be increasingly flexible and less labor intensive. Purely mechanical singulation installations, such as vibration feeders, no longer meet flexibility requirements or are no longer profitable, and manual work is becoming more expensive. One very cost-effective and flexible solution is the supply of parts in bulk,  as illustrated in Figure~\ref{fig:RBP} and Figure~\ref{fig:parts}, from which industrial robot arms pick out the objects one by one in order to feed them to the rest of the manufacturing chain. This application is referred to as random bin picking.  The goal is to locate one pickable object instance at a time and determine its six degree of freedom (6D) pose, so that the robot's end effector can be moved towards the object and grasp it.  This paper is focused on the object detection and localization task of such an industrial random bin picking application.



Pose estimation is a widely researched computer vision problem and various solutions based on 2D images, range images or 3D pointclouds exist.  However, very few are suited for the specific conditions of the real-world industrial bin picking application at hand.  As will be detailed in the next chapter's overview of relevant object detection algorithms, the lab examples mostly studied in literature differ quite a lot from real industrial bin picking scenarios.  In the latter case, objects with a wide gamut of characteristics are encountered: from nicely textured boxes to smooth shiny metal parts and from complex three dimensional free form shapes to rotationally-symmetric bolts. No satisfactory general solution to this problem exists yet.  In this paper, point pair features are studied as a versatile object detection technique in random bin picking. Moreover, combined with the simple but very powerful heuristic search space reduction that is proposed in this paper, the technique's computational demands remain within manageable bounds. We also propose a generic method that enables to use industrially available CAD models of the objects to be detected as input to our detection pipeline.

In literature, a hodgepodge of different evaluation mechanisms for random bin picking are used. This paper proposes a new universal way of evaluating pose estimation algorithms for random bin picking, necessary for fair comparison across different approaches. It consists of a completely automated procedure for the generation of realistic synthetic scenes and the evaluation of the detection algorithm.  As the procedure is automatic, it can be used in a closed loop to optimize the detection algorithms parameters. The goal is to achieve optimal performance across a set of widely varying objects.

The remainder of this paper is organized as follows. Section~\ref{RelatedWork} gives an extensive overview of the different 3D object localization techniques for random bin picking that are proposed in literature, as well as the detection evaluation methods available. Our point pair feature-based random bin picking approach is introduced in section~\ref{Methods}, composed by the dataset preprocessing, object detection, heuristic search space reduction and evaluation steps we propose.  Experimental results on representative industrial objects are presented and discussed in section~\ref{Results}, and the conclusions follow in section~\ref{Conclusions}.

\section{Related Work}
\label{RelatedWork}
An extensive review of the state of the art in 3D object detection and pose estimation algorithms is provided.  This section is split into a part discussing the point pair feature based techniques and a part discussing algorithms based on other representations.

\subsection{3D Object Detection Methods}
\label{3DDet}
In some simple cases the problem of detecting and estimating the pose of objects can be addressed by segmenting the scene and applying a global feature descriptor to the segmented parts in order to recognize one of the segments as being the considered object.  However, in our random bin picking case, this approach will not work, as it is not possible to reliably segment an object in the presence of significant clutter and occlusion.

A more sophisticated approach is to detect and describe certain parts or features of the object.  Some techniques have been proposed to detect 3D objects from regular images based on representations such as: 2D keypoints \cite{lowe2001local}, 2D edge templates\cite{ulrich2009cad}\cite{liu2012fast} or line descriptors\cite{damen2012real} \cite{tombari2013bold}.  Other techniques work on 3D representations such as: shape primitives (planes, cylinder, sphere, superquadrics, etc.)\cite{rabbani2006segmentation}, 3D keypoints\cite{tombari2010object}\cite{tombari2013performance}\cite{mian2010repeatability}, range image templates\cite{park2010fast} or color gradient and normal templates\cite{hinterstoisser2012gradient}\cite{BMVC2015_36}\cite{rios2013discriminatively}.

All these detection methods either create templates from several object views, or extract some kind of features.  An important downside to the methods relying on multiple object views is that they require a large amount of dense templates and as such, are computationally expensive to match.  There are two important issues with feature based methods, the first is that they are not very general: they can only represent objects that contain the specific type of feature they use.  Another issue is that the feature descriptors (e.g. 3D keypoint descriptors) are quite sensitive to noise.

\begin{figure*}
\centering
\subfloat[Depth image (dark colors are closer)]
{
	\includegraphics[width=0.45\textwidth]{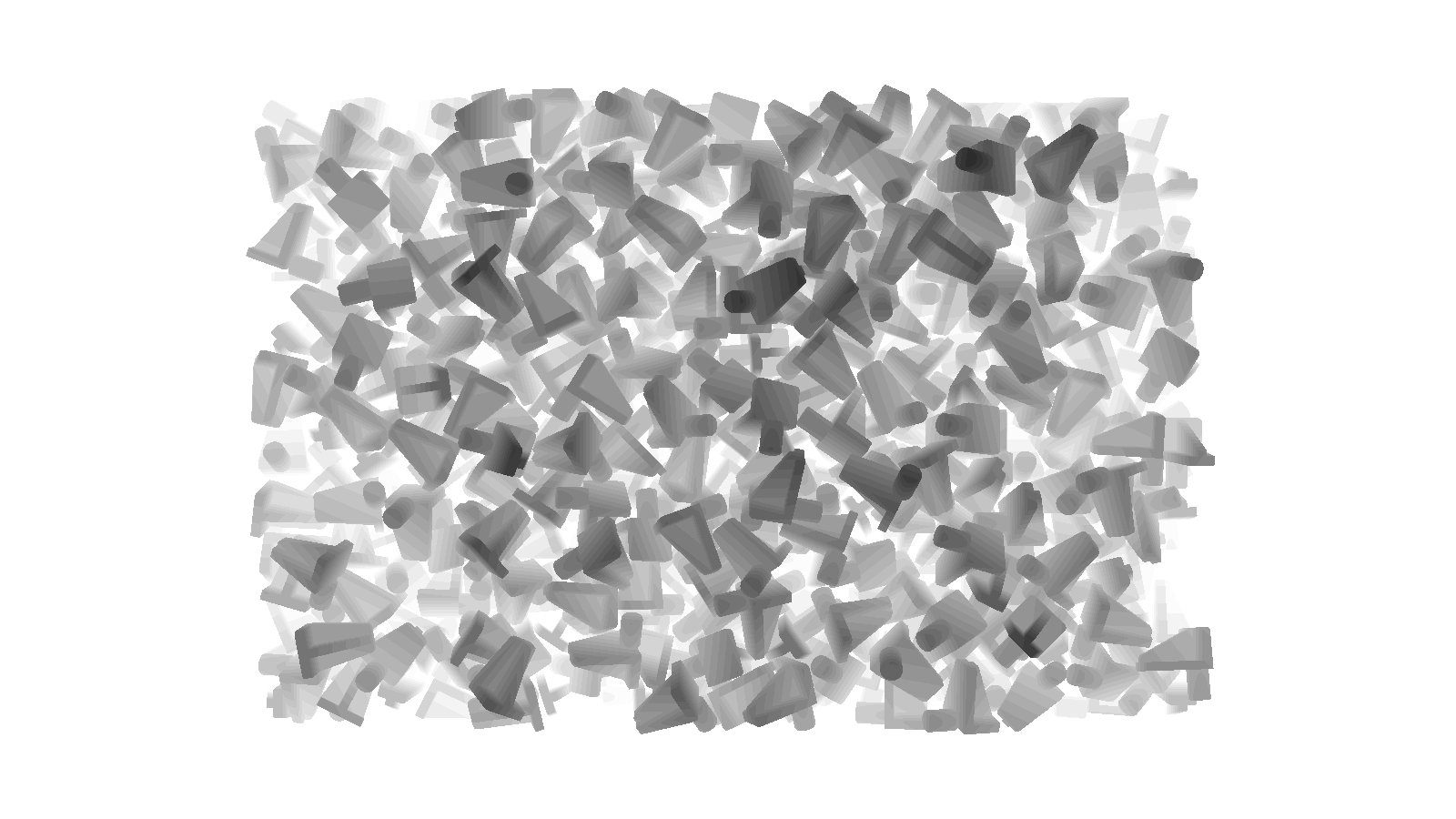}
	\label{fig:Depth}
}\hfill
\subfloat[Intensity image]
{
	\includegraphics[width=0.45\textwidth]{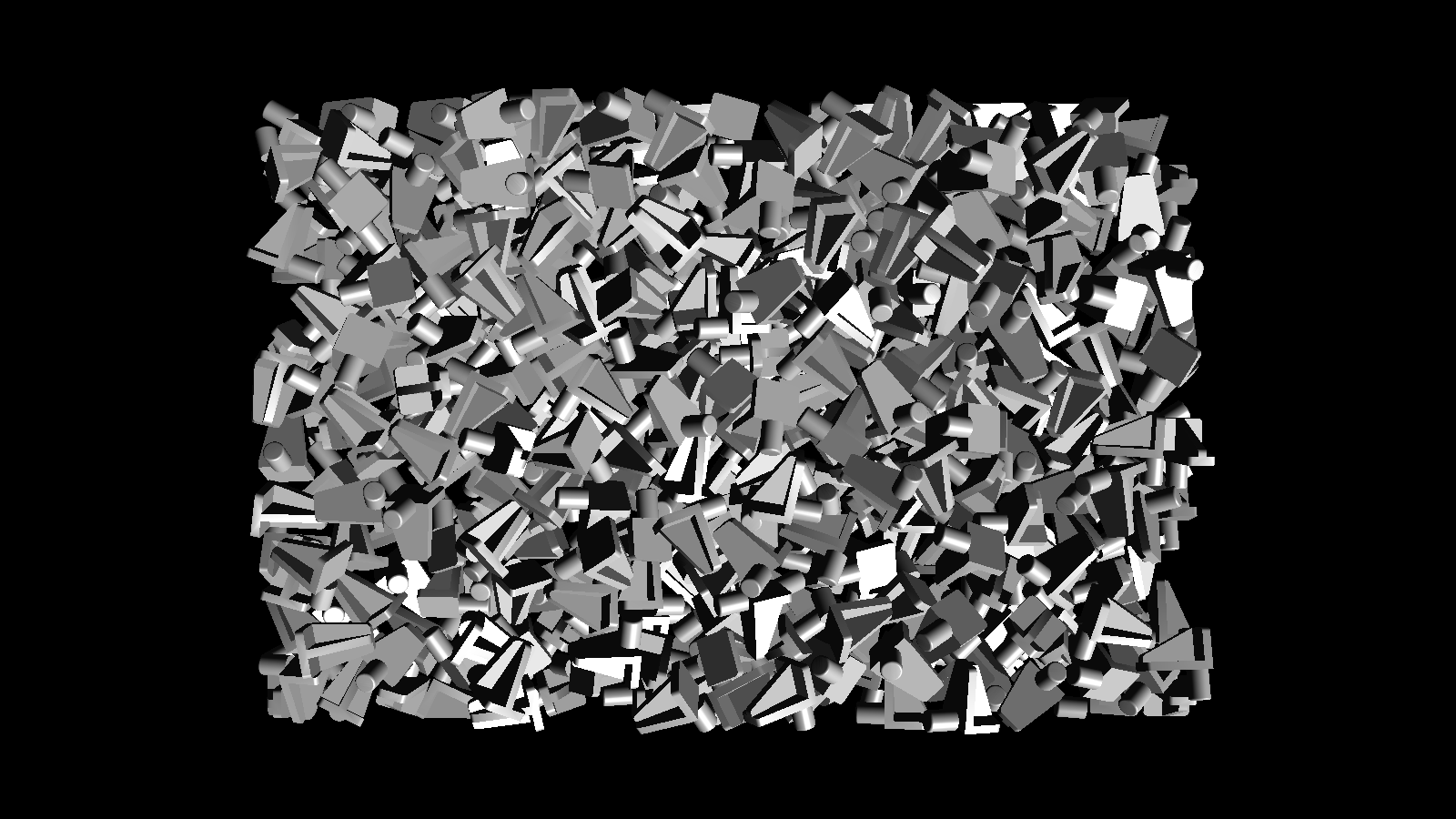}
	\label{fig:Intensity}
}
\caption{An example of a bin containing parts, as simulated by the proposed method and used in the experiments.  The bin itself was not rendered to allow better visibility, but was used in the physics simulation.}
\label{fig:Render}
\end{figure*}

\begin{figure*}
\centering
\subfloat[Input CAD model (644 vertices)]
{
	\includegraphics[width=0.28\textwidth]{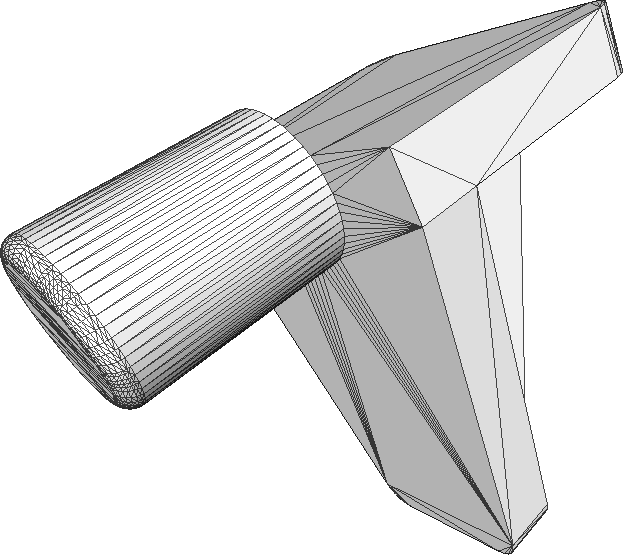}
	\label{fig:ObjectModel}
}\hfill
\subfloat[Object model decomposed into five convex parts using Approximate Convex Decomposition\cite{mamou2009simple}.  (185 vertices)]
{
	\includegraphics[width=0.28\textwidth]{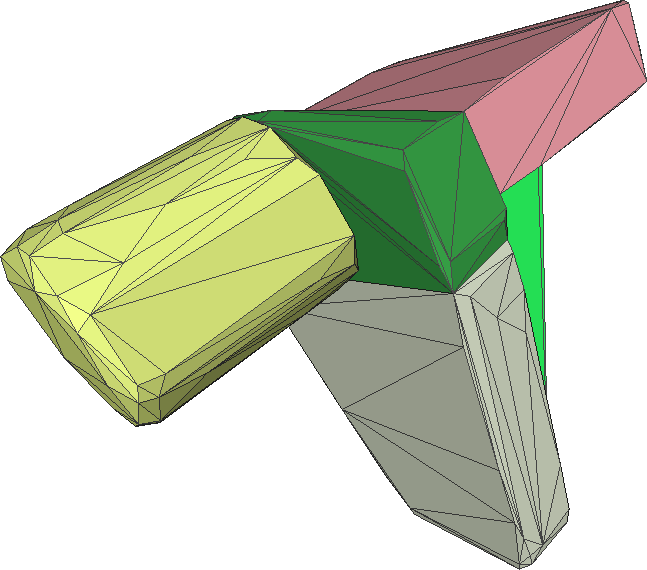}
	\label{fig:Convex decomposition}
}\hfill
\subfloat[Object model re-meshed using Approximated Centroidal Voronoi Diagrams\cite{valette2004approximated}\cite{valette2008generic} to use as input for the generation of a PPF model.   (1992 vertices)]
{
	\includegraphics[width=0.28\textwidth]{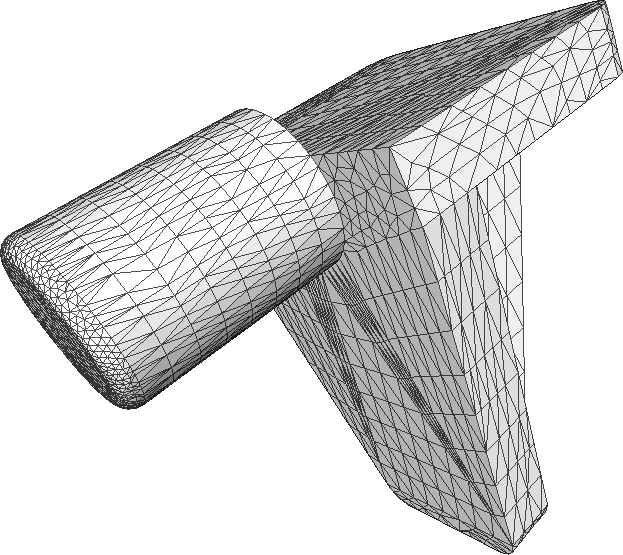}
	\label{fig:remesh}
}
\caption{The used 3D model and the derived models used in the simulation, training and detection steps.}
\label{fig:Model}
\end{figure*}

\subsection{Point Pair Features}
\label{PPF}

In the previous section methods relying on several types of features were discussed.  A lot of early work focused on describing the relations between a set of features of an object.  The feature vector used in point pair features (see Section~\ref{detection}) is very similar to some of these formulations\cite{stahs1992}, however, the important difference is that point pair features are calculated for all point pairs on a mesh, not just for specific features.  This means they can be used to represent free form objects.

Point pair features were first introduced for 3D object classification\cite{wahl2003surflet}.  A histogram of the features occurring on a surface allowed to classify the object.  This technique was later combined with a clustering algorithm to allow the detection of objects in pointclouds\cite{wahl2005cluster}.  An efficient procedure for storing the features in hashtables and calculating matches was proposed by Winkelbach et al.\cite{winkelbach2006low}, which was later extended to an object detection technique\cite{buchholz2010ransam}\cite{buchholz2013efficient}.

An efficient voting scheme to determine the object pose was described by Drost et al.\cite{drost2010model}.  The same authors later proposed a modified point pair feature using only points paired with geometric edge pixels, which are extracted from multimodal data\cite{drost2012multimodal}.  This makes the method suitable for object detection (instead of recognition).

Papazov et al.\cite{papazov2010efficient}\cite{papazov2012rigid} reduced the dimensions of the hashtable and the number of points to match by only using point pairs within a certain relative distance range from each other.  Kim et al.\cite{kim2011visibility} extended the point pair feature with a visibility context, in order to achieve a more descriptive feature vector.  Choi et al. proposed the use of more characteristic and selective features, such as boundary points and line segments\cite{choi2012voting} and extended the feature point vector with a color feature\cite{choi2012rgbd}\cite{choi2016rgbd}.  A method to learn optimal feature weights was proposed by Tuzel et al.\cite{tuzel2014learning}.  Several extensions are proposed by Birdal et al.\cite{birdal2015}: they introduce a segmentation method, add a hypothesis verification stage and weigh the voting based on visibility.

\subsection{Detection Evaluation}
\label{LitEvaluation}
Several ways to measure the accuracy of detections have been suggested in literature.
\begin{itemize}

\item The Average Distance between all vertices in the 3D model in the estimated pose and the ground truth pose\cite{hinterstoisser2012model} \cite{papazov2010efficient} \cite{gelfand2005robust}.  In some cases this corresponds to the residual error reported by the Iterative Closest Point (ICP) algorithm used to refine the pose.
\item The fraction of points of the object model points transformed to the detected pose that are within a predefined distance to their closest point\cite{buchholz2010ransam}.
\item The Intersection Over Union of the 2D axis aligned bounding boxes of the estimated and ground truth pose.  This method could be used to compare the performance to algorithms that only provide a 2D position estimate.  However, it obviously provides very little information on the depth and rotation accuracy. \cite{everingham2005pascal}\cite{damen2012real}
\item Picking success rate: the number of times a part is successfully grasped and extracted from the bin by the robot divided by the total number of attempts.  This method requires implementation of the system using an actual sensor and robot.  It can be argued that it is the most representative evaluation method to evaluate the actual system performance.  An important advantage is that it allows to evaluate performance without requiring ground truth object pose data.  The downside however, is that it is quite difficult to asses the detection algorithm performance, as variations in performance may be due to other parts of the system.  It would also require users to have the exact same system, should they wish to compare their methods.  \cite{buchholz2010ransam}\cite{choi2012voting}
\item The rotation and translation error, in absolute or relative measures. \cite{drost2010model}\cite{choi2012voting}\cite{tuzel2014learning}\cite{birdal2015}
\end{itemize}

\section{Methods}
\label{Methods}

\subsection{Generating Synthetic Random Bin Picking Datasets}
\label{Dataset}
To obtain realistic test data that are representative for random bin picking, a method was developed to automatically generate synthetic random bin picking scenes from 3D object models.  The 3D models used as input can be either the original CAD data, or a reverse-engineered model from a combination of 3D scans of the object.  A 3D model of a bin is loaded and a set of parts is dropped into it.  The bullet physics library\cite{coumans2015} is used to detect inter-object collisions and to calculate the resulting forces and dynamics.  To achieve acceptable simulation times for the large number of objects being simulated, the objects are first decomposed into convex parts using Approximate Convex Decomposition\cite{mamou2009simple} as shown in Figure~\ref{fig:Convex decomposition}.  The underlying reason for the speed gain is the fast methods that can be used to check for collisions between convex objects, versus the slower collision checking methods for general, free-form meshes.

\begin{figure}
\centering
\includegraphics[width=0.48\textwidth]{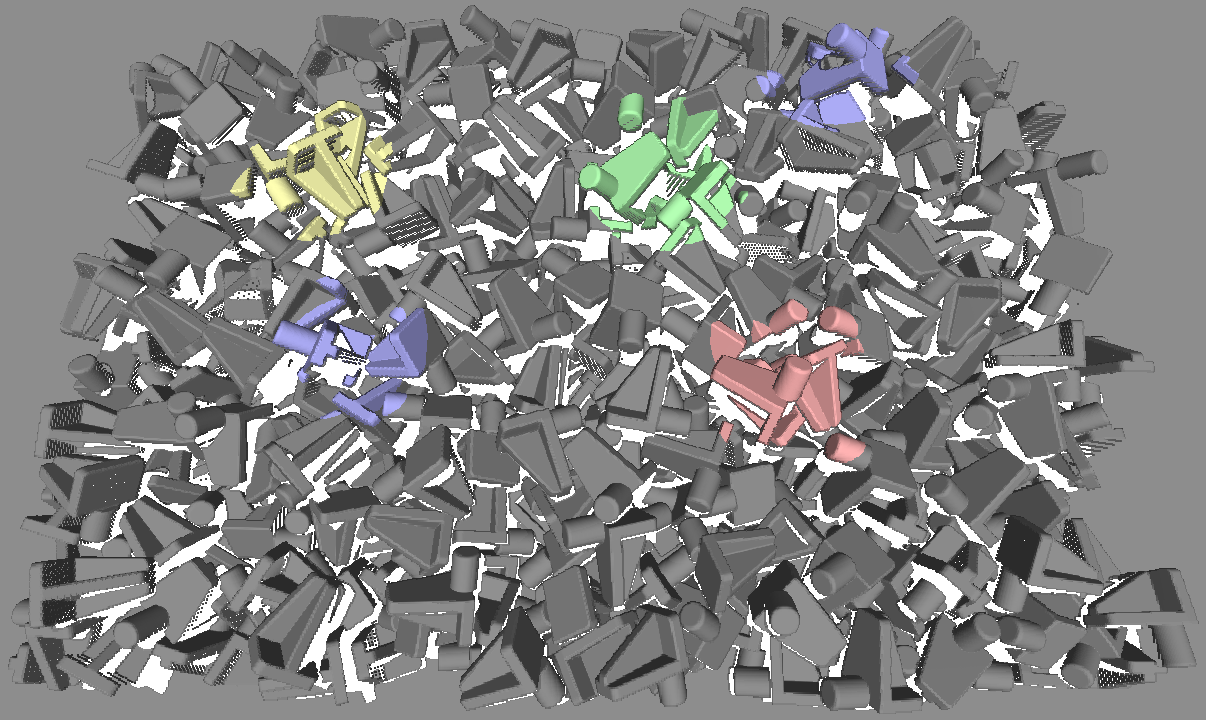}
\caption{A view of the 3D mesh of the generated scene.  The local neighborhoods of the five highest points in the scene are highlighted in color.  When multiple hypotheses are tested, we propose picking the one that accumulated the highest number of votes in the detection process.  To ensure better visibility, the gray scene points were very slightly translated downwards.}
\label{fig:highpoints}
\end{figure}

Once all objects have reached a stable situation, meaning they are no longer in motion, their 6D poses are stored as the ground truth.  To generate the corresponding range and intensity image (Figure~\ref{fig:Depth} and~\ref{fig:Intensity}) and the surface mesh (Figure~\ref{fig:highpoints}), the approximated models are replaced with the original models (Figure~\ref{fig:ObjectModel}) and the scene is rendered from the sensor's point of view.

The rendering can be performed very rapidly on the GPU, which automatically handles issues such as the object transformations, determining visibility and interpolation between model points.  The point coordinates can be calculated from the GPU's z-buffer values as:

\renewcommand{\arraystretch}{2.5}
\begin{equation}
   \left[\begin{array}{c}
         x \\
	y \\
	z
    \end{array}\right]
_{(u,v)}
 =    \left[\begin{array}{c}
z_{(u,v)}\dfrac{u}{f} \\
z_{(u,v)}\dfrac{v}{f} \\
\dfrac{d_{n}d_{f}}{d_{f}+b_{(u,v)}(d_{n}-d_{f})}
    \end{array}\right]
\end{equation}

With $(u,v)$ the pixel coordinates relative to the center of the image,  $f$ the focal length, $d_{n}$ and $d_{f}$ the near and far clipping plane distance and $b_{(u,v)}$ the z-buffer value for a pixel $(u,v)$.

\subsection{Point Pair Feature based Object Detection}
\label{detection}
\begin{figure}
\centering
\includegraphics[width=0.3\textwidth]{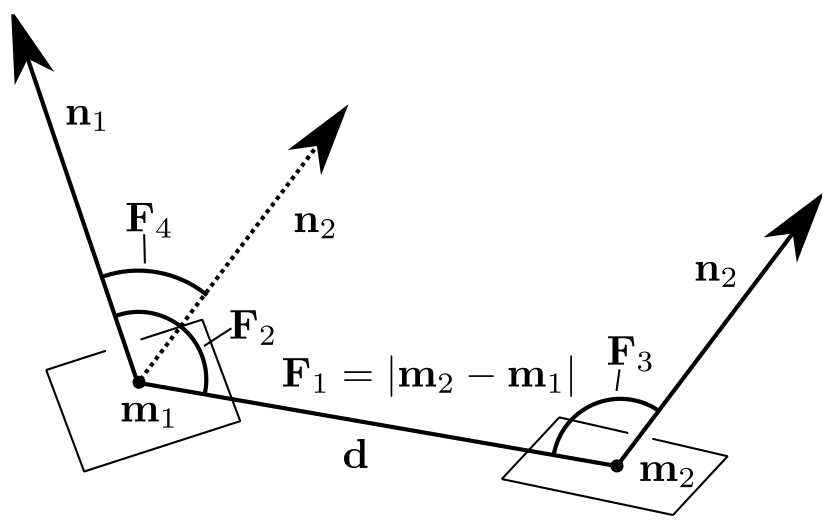}
\caption{Two surface points $\mathbf{m_{i}}$ and their normals $\mathbf{n_{i}}$ determine a point pair feature.}
\label{fig:ppf}
\end{figure}

Point pair features describe the relative position and orientation of points on the surface of an object (Figure~\ref{fig:ppf}).  Drost et al. define the feature\cite{drost2010model} as: For two points $\mathbf{m_{1}}$ and $\mathbf{m_{2}}$ with normals $\mathbf{n_{1}}$ and $\mathbf{n_{2}}$ , we set $\mathbf{d} =  \mathbf{m_{2}} - \mathbf{m_{1}}$ and define the feature  $\mathbf{F}$ as:

\begin{equation}
\mathbf{F}(\mathbf{m_{1}} , \mathbf{m_{2}} ) = ( \lVert \mathbf{d} \rVert _{2} , \angle(\mathbf{n_{1}} , \mathbf{d}), \angle(\mathbf{n_{2}} , \mathbf{d}), \angle(\mathbf{n_{1}} , \mathbf{n_{2}} )) 
\end{equation}

where $\angle(\mathbf{a}, \mathbf{b}) \in [0 \quad \pi]$ denotes the angle between two vectors.

To create a new object model, the four dimensional feature vectors are discretized and stored in a hashtable.  When using a CAD model as input, some model preprocessing steps are usually needed, as CAD models typically have very non-uniform meshes (Figure~\ref{fig:ObjectModel}).  We apply a re-meshing technique based on Approximated Centroidal Voronoi Diagrams\cite{valette2004approximated}\cite{valette2008generic} to obtain a much more uniform mesh with a suitable resolution (Figure~\ref{fig:remesh}).  Note that this method works well both for input models that have too little vertices, or too many vertices.  This method also offers an improvement in usability compared to rendering the object from a large number of viewpoints, calculating the resulting pointcloud and then stitching all generated pointclouds into a new object model.  After re-meshing, a voxel based uniform subsampling procedure is used to retain the desired number of points.

The same detection procedure as Drost et al.\cite{drost2010model} is used. To detect the object model in the scene, the uniform subsampling procedure is applied to the scene mesh.  An additional subsampling procedure is used to select reference points among the scene points.  The point pair features between these reference points and all other subsampled scene points are calculated, and their feature vector is compared to the ones stored in the object models hashtables.  All corresponding features are obtained from the model and vote for an object pose.  In a pose clustering procedure, similar poses are grouped and their poses are averaged.

\begin{figure*}
\centering
\subfloat
{
	\includegraphics[width=0.3\textwidth]{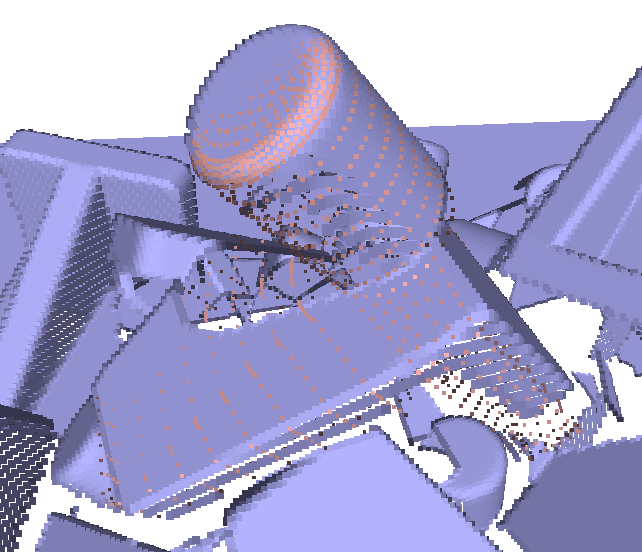}
	\label{fig:detection1}
}\qquad \qquad \qquad \qquad \qquad
\subfloat
{
	\includegraphics[width=0.3\textwidth]{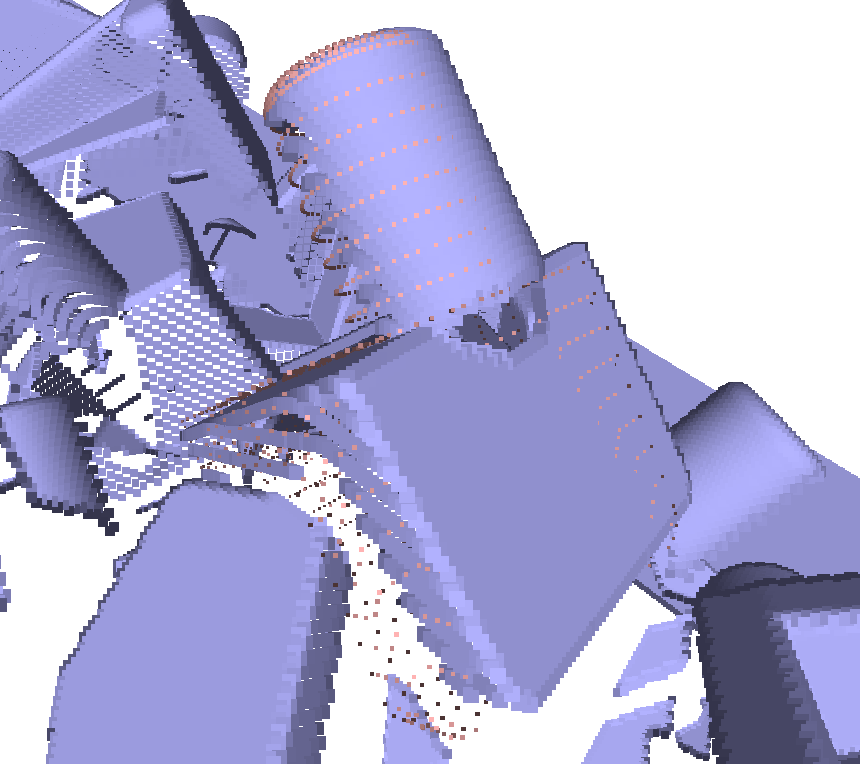}
	\label{fig:detection2}
}
\caption{A detected object model (pink points) in the scene pointcloud (blue points),  as seen from two viewpoints.}
\label{fid:Detection}
\end{figure*}

\begin{figure*}
\centering

\subfloat[Varying translation threshold (all detections)]
{
	\includegraphics[width=0.48\textwidth]{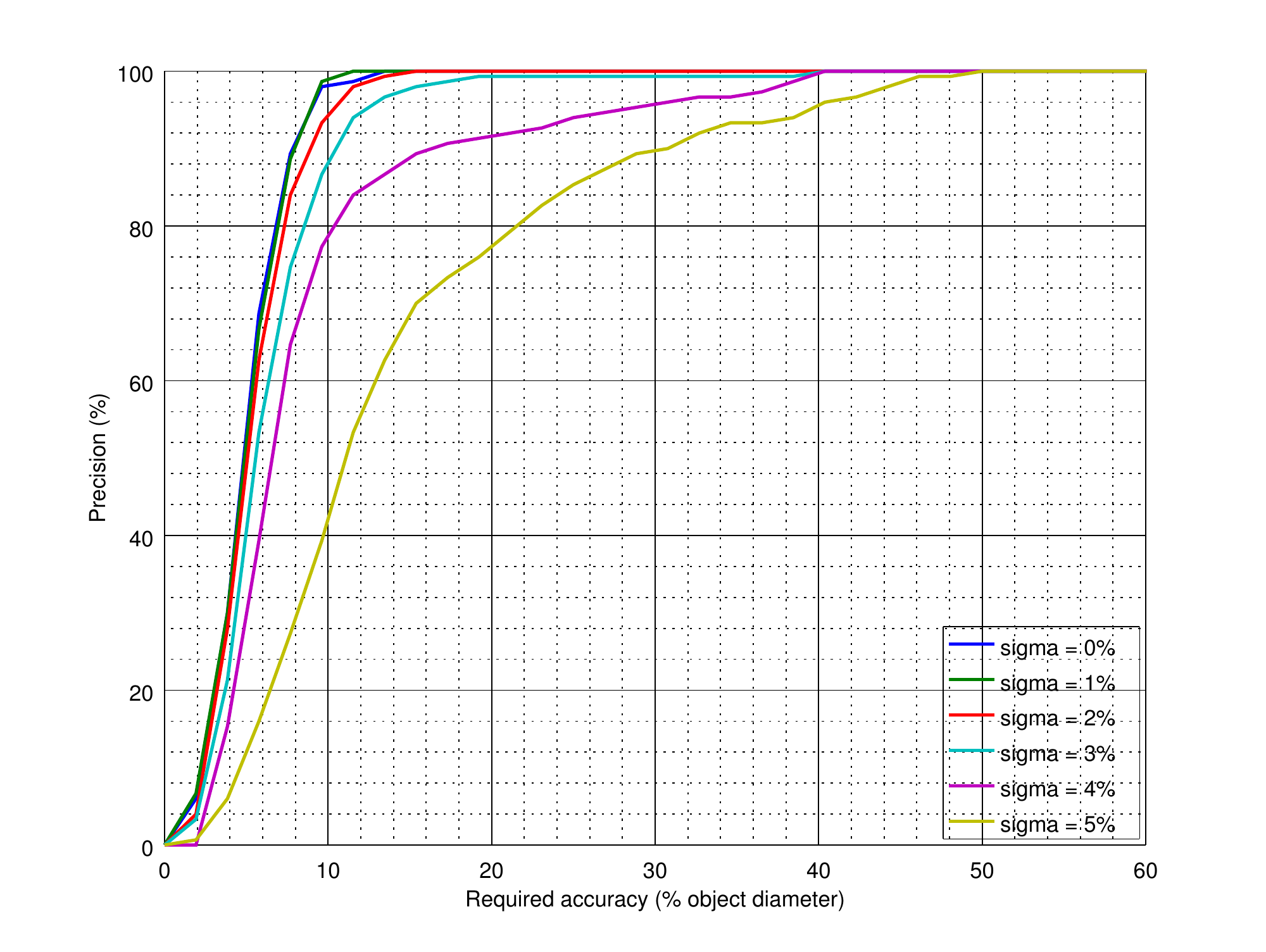}
	\label{fig:MeanTransAcc}
}
\hfill
\subfloat[Varying translation threshold (detections with highest number of votes)]
{
	\includegraphics[width=0.48\textwidth]{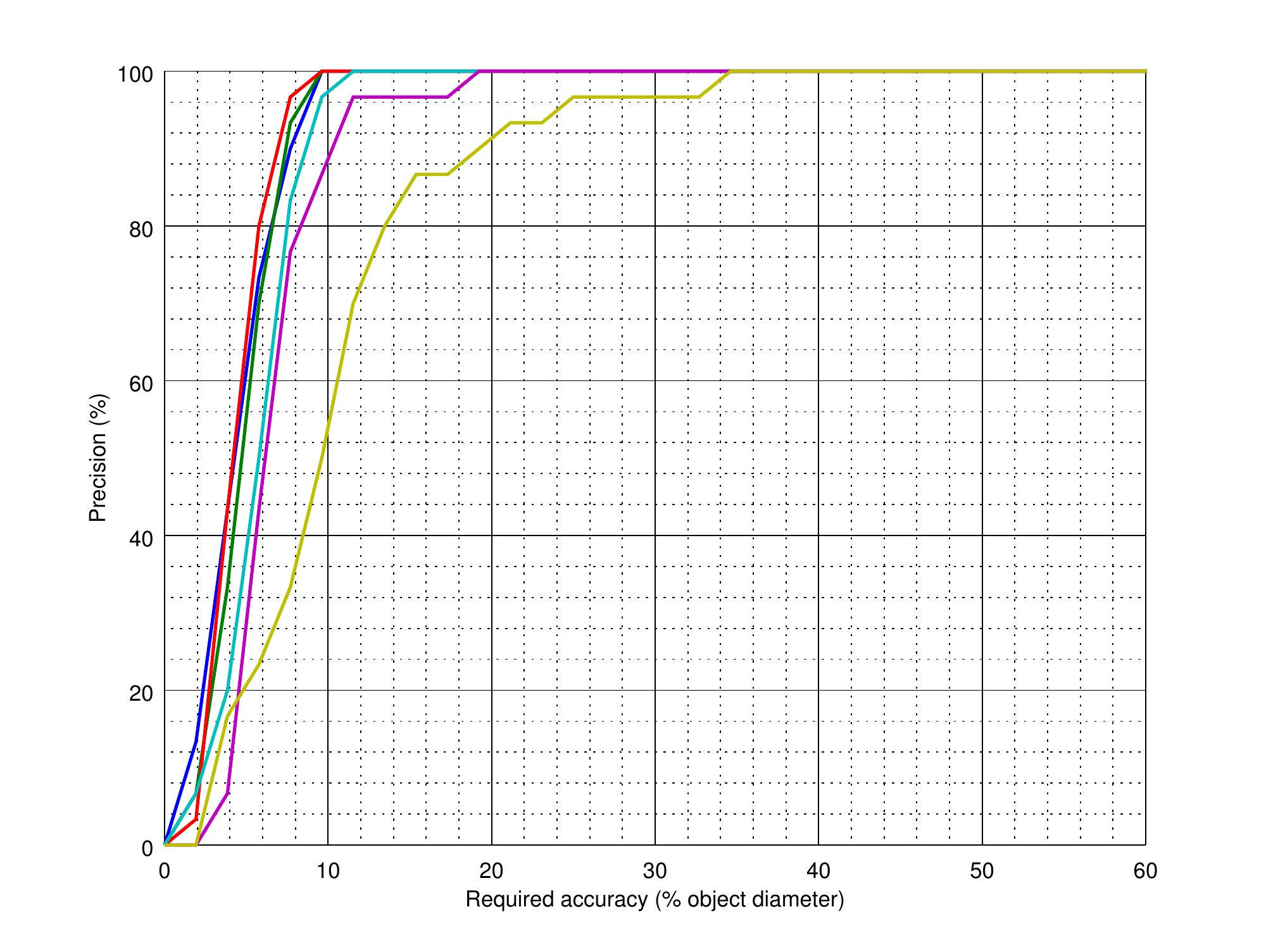}
	\label{fig:MaxvTransAcc}
}
\\
\subfloat[Varying rotation threshold (all detections)]
{ 
	\includegraphics[width=0.48\textwidth]{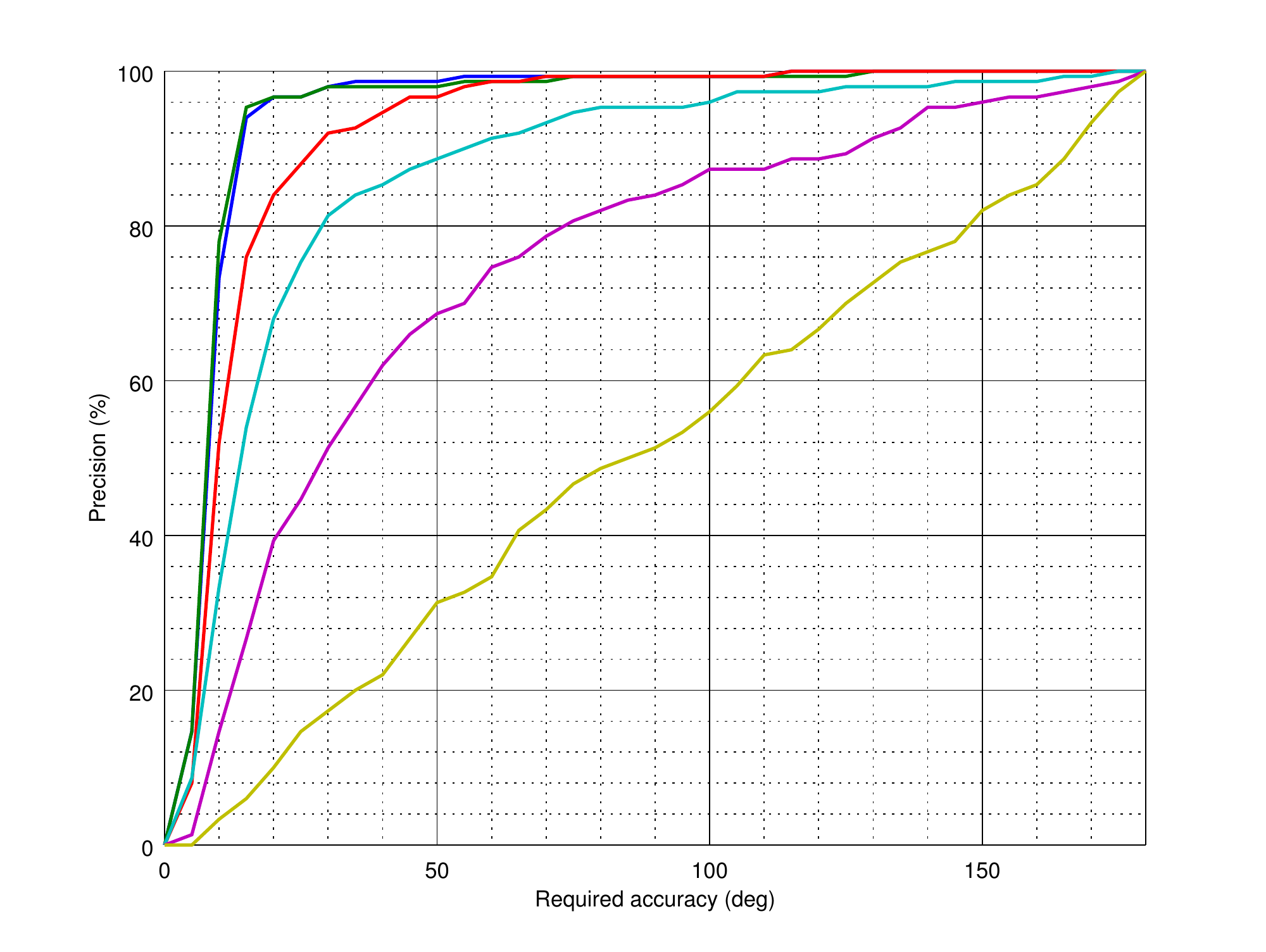}
	\label{fig:MeanRotAcc}
}
\hfill
\subfloat[Varying rotation threshold (detections with highest number of votes)]
{ 
	\includegraphics[width=0.48\textwidth]{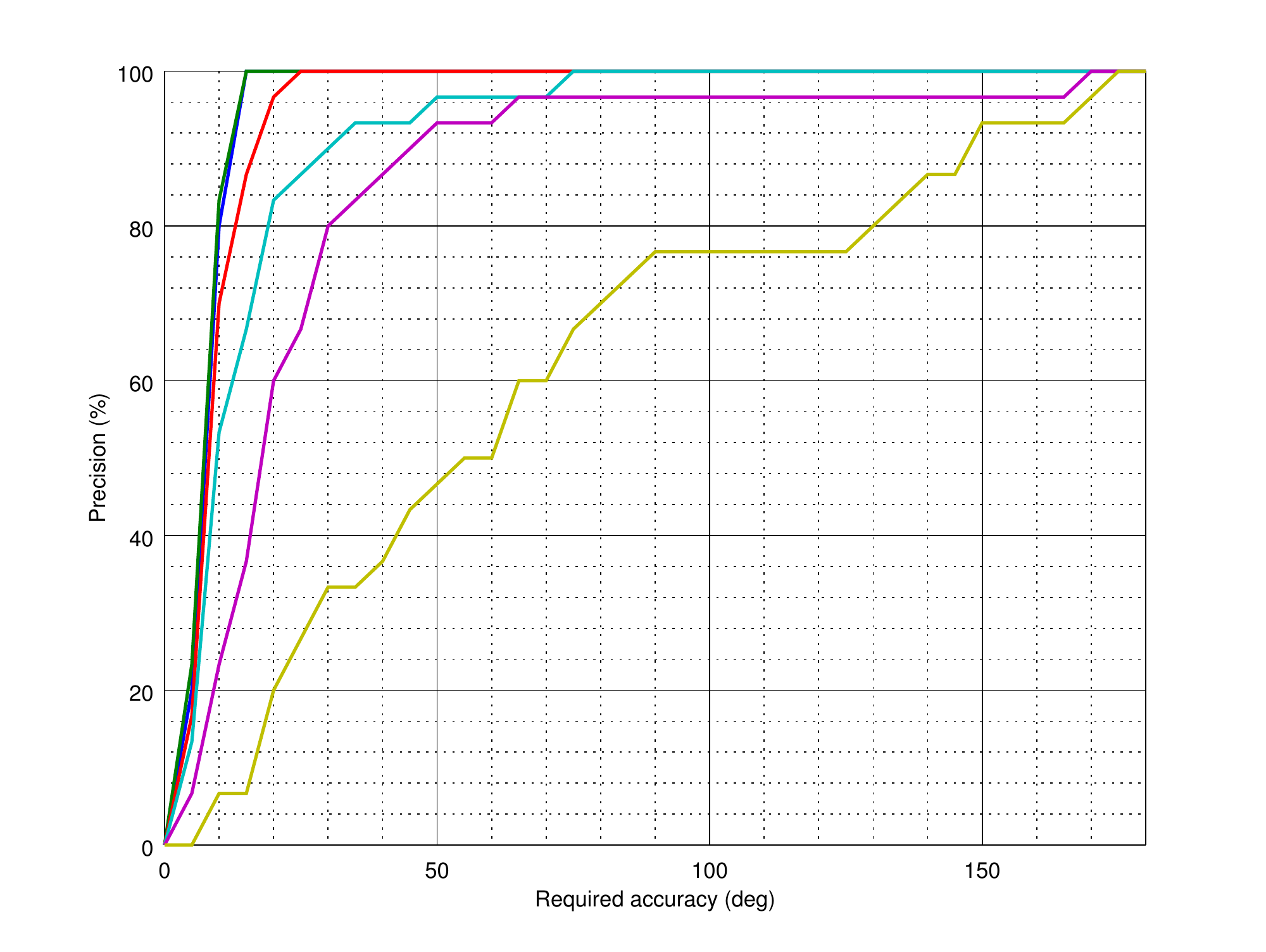}
	\label{fig:MaxvRotAcc}
}
\caption{Detection precision as a function of increasingly tight error thresholds and an increasing amount of gaussian noise (color coded, with $\sigma$ 0 to 5\% of the object diameter).}
\label{fig:Precision}
\end{figure*}

\subsection{Heuristic Hypothesis Generation}
\label{hypothesis}
Due to the high computational complexity of detecting objects using point pair features, it is not feasible to detect small objects in large pointclouds directly.  Therefore, a heuristic approach is used to to drastically reduce complexity.  A low pass filter is applied to the range image, and iteratively (five iterations in our application), the highest point is selected.  After each iteration, the local neighborhood of the selected point is excluded.    This is done by putting a threshold on the distance between the points and the highest point being used.  The radius of the regions is chosen to be slightly larger then the object's diameter.

The assumption made is that the highest objects will be the easiest to detect and grasp, as they are less likely to be occluded by other objects.  This method does not rely on segmentation (e.g. \cite{birdal2015}) or feature (boundary) detection (e.g. \cite{buchholz2013efficient}), which would strongly limit the applicability, as they depend on features being present and detectable.  An example of the hypothesized object positions is shown in Figure~\ref{fig:highpoints}.

\subsection{Evaluation}
\label{Evaluation}
The detection algorithm was evaluated on the synthetic scenes generated from a 3D object model.  The detection is performed in the local neighborhood of the hypothesized positions.  For each hypothesis, the cluster with the highest number of votes is selected, the contributing votes are averaged and the resulting pose is compared to the ground truth. 

The translation error is defined as the euclidean distance between the detection and ground truth pose.  The translational error can be expressed as a percentage of the object model diameter to allow comparison among objects of different sizes.  The rotational error is obtained by calculating the rotation matrix to align the frames of the detection and ground truth pose.  The rotation matrix is converted to Angle-Axis representation and the resulting angle is used as an error metric.  This means the rotation error is in the range $[0 \quad \pi]$.  

As the entire procedure of scene synthesis, model creation, object detection and evaluation is automatic, it can be used in a closed loop to optimize the detection algorithm parameters.

\section{Experimental Results}
\label{Results}
A synthetic dataset of 30 bins is generated to test the point pair feature based object detection.  The objects are dropped in an array arrangement of five by seven, at a time.  This is repeated ten times per bin, so each bin contains 350 objects.  The initial object orientation is random, and the initial velocity is zero.  The objects fall into the bin as a result of the applied gravitational force.  The rendering resolution was set to 1600x900 pixel.  The used object has a diameter of about 5.2 and the bin dimensions are 60x40x30 (width x height x depth).  The objects are dropped into a bin from a height of 60 to the bin floor, while the camera is positioned at 100 from the bin floor.  

The five highest points in the scene were selected as hypotheses to run the detection algorithm on.  The performance of the detection with respect to increasing levels of gaussian noise was tested (Figure~\ref{fig:Precision}).  Note that no effort was made to reliably estimate the surface normals in the presence of noise, as this is not the key issue addressed in this paper.  Adding such a method would drastically reduce the negative influence of the applied noise.

The point pair features were discretized into 30 steps for the normal angles, resulting in an angle of 12$^{\circ}$ per step.  The distance between the point pairs was discretized into 20 steps, and the maximum distance corresponds to the object diameter.  The resolution of the subsampling applied to both the model and scene was set to five percent of the object diameter.  The experiments were performed using 20\% of the subsampled scene points as reference points.  All poses within a distance of 0.75 and with a rotation angle of 20 degrees or less are clustered.  Their poses are averaged to increase the precision of the final pose estimate.  Using these settings, the average detection time per hypothesis is 726 milliseconds.  Note that the percentage of reference points can be reduced to increase speed, while still achieving high performance\cite{drost2010model}.  When 2\% of the subsampled points are used as reference points, the required processing time is only 70 milliseconds per hypothesis.

The precision of using any of the five hypotheses was compared to using the one with the highest number of supporting points (points that vote for that particular pose).  The improvements can be seen by comparing Figures~\ref{fig:MeanTransAcc} and \ref{fig:MeanRotAcc} to Figures~\ref{fig:MaxvTransAcc} and \ref{fig:MaxvRotAcc}.  The results indicate that the number of votes received from supporting points is a valuable indicator of whether a detection is correct.

\section{Conclusions}
\label{Conclusions}
An object detection and pose estimation algorithm based on point pair features was proposed.  Several enhancements to optimize performance in a random bin picking context were made.  Experiments on synthetic data show that we can drastically reduce the complexity of the problem by reducing the search space using simple heuristics.  This offers huge benefits, as it increases robustness, speed and accuracy.  The accuracy of the estimated poses was measured and found to be sufficient for our purpose, without requiring an additional ICP step.

\section*{Acknowledgment}
\label{Acknowledgment}
This work was partially funded by IWT (TETRA project RaPiDo, \#140358).  We thank our project partner ACRO and the participating companies.  Special thanks to Wiebe Van Ranst for helping track down some hard to find bugs.

\bibliographystyle{IEEEtran}
\bibliography{IEEEabrv,WAB_CRV2016}

\end{document}